%% file: main.tex
\documentclass[a4paper]{article}
\usepackage[final]{nips_2019}

\usepackage{hyperref}
\usepackage{url}
\usepackage{natbib}
\usepackage{booktabs}
\usepackage{color}
\usepackage{placeins}
\usepackage{amsmath}
\usepackage{amsthm}
\usepackage{amssymb}
\usepackage{caption}
\usepackage{subcaption}
\usepackage{tikz}
\usepackage{comment}
\usepackage{bm}
\usepackage{array}
\usepackage{times}
\usepackage{scalefnt}
\usepackage{mathtools}
\usepackage{algorithm}
\usepackage{algorithmic}
\usepackage{footnote}
\usepackage{enumitem}

\usepackage[english]{babel}
\usepackage[utf8x]{inputenc}
\usepackage[T1]{fontenc}

\usepackage{amsmath}
\usepackage{amssymb}
\usepackage{graphicx}
\usepackage[colorinlistoftodos]{todonotes}

\usepackage{graphicx,caption}
\usepackage{enumitem}
\usepackage{psfrag}
\usepackage{amsmath,amssymb,epsf}
\usepackage{verbatim}
\usepackage{import}
\usepackage{mathtools}

\input{macro.tex}

\title{Approximating Human Judgment of \\Generated Image Quality}
 \author{Y. Alex Kolchinski\thanks{Equal contribution.}~, Sharon Zhou\footnotemark[1]~, Shengjia Zhao, \\ \textbf{Mitchell Gordon, Stefano Ermon} \\
 Stanford University\\
 \texttt{\{yakolch, sharonz, sjzhao, mgord, ermon\}@cs.stanford.edu} \\
 }

\begin{document}
\maketitle
\vspace{-3mm}
\begin{abstract}
    Generative models have made immense progress in recent years, particularly in their ability to generate high quality images. However, that quality has been difficult to evaluate rigorously, with evaluation dominated by heuristic approaches that do not correlate well with human judgment, such as the Inception Score and Fr\'echet Inception Distance. Real human labels have also been used in evaluation, but are inefficient and expensive to collect for each image. Here, we present a novel method to automatically evaluate images based on their quality as perceived by humans. By not only generating image embeddings from Inception network activations and comparing them to the activations for real images, of which other methods perform a variant, but also regressing the activation statistics to match gold standard human labels, we demonstrate 66\% accuracy in predicting human scores of image realism, matching the human inter-rater agreement rate. Our approach also generalizes across generative models, suggesting the potential for capturing a model-agnostic measure of image quality. We open source our dataset of human labels for the advancement of research and techniques in this area.
\end{abstract}

\section{Introduction}
Generative models have recently been used with great success for various tasks including the generation of highly realistic images \citep{brock2018large, karras2017progressive}. The growing capabilities of these models---now with applications in medicine~\citep{nie2017medical}, fashion~\citep{rostamzadeh2018fashion}, and music~\citep{engel2019gansynth}---represent a promising frontier, yet one pressing challenge has been their evaluation~\citep{barratt2018note, borji2019pros}. In particular, evaluating the realism of generated samples, which is useful for most applications, has ultimately relied on human judgment. Widely used metrics which calculate statistics on generated and real samples have been proposed for automatic inter-model comparison, like the Inception Score or IS \citep{salimans2016improved} and Fr\'echet Inception Distance or FID \citep{heusel2017gans}. However, these metrics only assess models holistically instead of providing sample-level scores of the realism of individual images. Moreover, they fail to correlate with perceptual realism, except at a coarse level~\citep{zhou2019hype}. As a result, studies typically resort to collecting human labels for individual images to judge the realism of a model's outputs. 

However, gathering human assessments of individual images' quality is a slow and difficult process. While recent work has made strides to reduce sample size and efficiently evaluate generative models using humans directly~\citep{zhou2019hype, hashimoto2019unifying}, this evaluation is similar to its automated counterparts in judging models' overall ability to produce realistic outputs and not sample-level realism. This is not surprising, as evaluation at the sample level is substantially more expensive, requiring at least one human to evaluate each image, and benefiting from multiple human labelers in order to be able to compare perceived quality between images and counteract biases that may occur from a single evaluator~\citep{gordon2013reporting}.

In response to the lack of efficient evaluation metrics that take into account human judgment, we introduce a novel automated method for scoring individual images on their perceptual realism. 
While previous methods manually design a single quality score that is loosely correlated with human realism, we train our model using both feature embeddings and human labels, generating multiple scores from different layers of the pretrained network and regressing these scores to match human perceptual scores of realism. We evaluate our model on several held-out test sets and achieve 58.8-66.4\% accuracy compared to human ground truth scores---this is compared to a 58.9\%-65.5\% inter-human agreement. In attaining this, we hope that our work advances research in relating automated metrics directly to human perception.

\section{Method}
\subsection{Human Realism Scores}
We run HYPE~\citep{zhou2019hype} to collect human labels of perceptual realism on 4803 images generated from different generative models \{ProGAN, StyleGAN\} trained on CelebA. The HYPE procedure produces binary labels of ``real" or ``generated" on each image. For each generative model, we randomly split the dataset of image-label pairs into training and test splits, where test sets have 10\% of the data. The exact numbers are \{1787, 2483\} in the training sets and \{223, 310\} in the test sets of ProGAN and StyleGAN, respectively.\footnote{For a copy of the dataset, please contact the authors.}

\subsection{Embeddings}
We construct embeddings for each image using activations from different layers of an ImageNet-pretrained Inception-V3 network. We selected layers for a range of low-level to high-level features and a varying degree of regularization from the auxiliary classifier.


We first consider a single layer. Let $f: \mathcal{X} \to \mathbb{R}^{W \times H \times C}$ be the inception network that maps an image $x \in \mathcal{X}$ to its activation on that layer $z = f(x)$, which is a $W\times H \times C$ tensor. 

Given a set of $n$ training images $\lbrace x^{(1)}, \cdots, x^{(n)} \rbrace$ we can compute activations for each of them $\lbrace z^{(1)} = f(x^{(1)}), \cdots, z^{(n)} = f(x^{(n)}) \rbrace$ as our set of reference tensors. Intuitively, if an image $x^{\mathrm{test}}$ is similar to real images, then its activation $z^{\mathrm{test}} = f(x^{\mathrm{test}})$ should also be similar to some activation in the reference set. More formally, for each $1 \leq u \leq W$ and $1 \leq v \leq H$ we compare the $C$-dimensional vector  $z^{\mathrm{test}}_{uv} \in \mathbb{R}^C$ to the set of vectors in the same spatial location $\lbrace z^{(1)}_{uv}, \cdots, z^{(n)}_{uv} \rbrace$, and we compute the distance to the nearest neighbor 
\begin{align*}  
d_{uv}^{\mathrm{test}} &= \mathrm{NearestNeighborDistance}(z^{\mathrm{test}}_{uv}, \lbrace z^{(1)}_{uv}, \cdots, z^{(n)}_{uv} \rbrace) =
\min_{i=1,\cdots,n} \lVert z^{\mathrm{test}}_{uv} - z^{(i)}_{uv} \rVert_2
\end{align*}
Finally we aggregate these vectors $d^{\mathrm{test}} = \sum_{1 \leq u \leq W, 1 \leq v \leq M} d^{\mathrm{test}}_{uv}$ into a single real number. 

We can perform the above procedure for different layers of the Inception-V3 network. Specifically we choose layers \textit{\{Conv2d\_1a\_3x, Conv2d\_2b\_3x3, Conv2d\_3b\_1x1, Mixed\_5d, 
Mixed\_6e, Mixed\_7c, FC\}} from Inception-V3. We treat the number of layers $m$, here seven, as a hyperparameter that trades off expressiveness and efficiency; $m=1$ on the last pooling layer reduces to activations for FID. We also treat fully connected layers as a special case where $W=H=1$. Because each layer generates a single real number $d^{\mathrm{test}}$, using $m$ layers produces $m$ real numbers for each test image $x^{\mathrm{test}}$. We then use these real numbers to regress the human perception score. Intuitively, these real numbers correspond to how similar the activation patterns of the input image are to real images for layers representing low-, mid-, and high- level features. To accelerate computation, we subsample 10,000 vectors per layer to serve as a representative sample of the activation patterns in lieu of computing distances between full layers.

\subsection{Model Fitting}
We fit a logistic regression model to predict the binary (real/fake) human labels based on the reference set. We choose a low-dimensional data representation in combination with a linear model to best demonstrate the ability of the embedding distances to predict human perception, using limited training data. Data on human labels is limited due to the cost of large-scale annotation, though if abundant, could enable higher-dimensional representations. We evaluate the efficacy of the logistic regression model on held-out test sets for each generative model, as well as across data sets to determine how well the logistic regression model trained on ProGAN data generalizes to StyleGAN, and vice versa. 



\section{Evaluation and Results}
We employ two methods of evaluation to compare our model's judgments against human perception. The first evaluates a model's accuracy at predicting human labels of previously unseen images, both within and across data sets. For these binary test sets, we report percent agreement of each model's human labels, within and across dataset.  The second evaluates how closely correlated the models' continuous quality scores are with the fractions of humans who judge an image as real. On these realism spectrum test sets, we compute Spearman's rank-order correlation coefficient $\rho$ between the model's outputs and mean human scores per image. See Table~\ref{table:1}.
Our model's accuracy scores indicate promise, but are clearly not sufficient for deployment as-is. Significant factors limiting the performance of the model in this experiment are the inconsistency of human labels and the small size of the data set. Both would be addressed by a larger dataset.

\begin{table}[ht]
\centering

\caption{The least (leftmost 3) and most (rightmost 3) realistic BeGAN images, as judged by one of the best-performing models.}
\begin{tabular}{ccccccc}
\includegraphics[width = .6in]{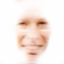} &
\includegraphics[width = .6in]{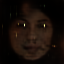} &
\includegraphics[width = .6in]{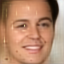} &
&
\includegraphics[width = .6in]{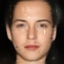} &
\includegraphics[width = .6in]{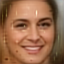} &
\includegraphics[width = .6in]{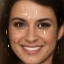}
\end{tabular}
\end{table}

\subsection{Binary Test Sets}

We begin by fitting a model to each of the two training sets, which contain binary human judgments for whether each image in the training set is real or fake. In the ProGAN dataset, there are 1456 ``real'' and 1530 ``fake'' labels, a fairly even split. StyleGAN skews more realistic, with 2983 ``real'' and ``1814'' fake human labels.  We then test each model on the in-distribution test set, e.g. a StyleGAN-trained model evaluated on the StyleGAN test set. We also test each of the models on the out-of-distribution test set to evaluate how well the automatic metrics of image realism generalize across datasets. The scores are the percentage of human labels which are correctly predicted by the relevant logistic regression model thresholded at $50$\%.

\textbf{Results.} Our models predict roughly two out of three human labels correctly when tested in-domain. Notably, the out-of-domain performance is comparable, with the ProGAN model achieving 65\% accuracy when tested on StyleGAN data, and the StyleGAN model tested on ProGAN data not far behind at 59\% accuracy. In fact, inter-human agreement is far from perfect, at 65.5\% for ProGAN and 58.9\% for StyleGAN. 

\subsection{Realism Spectrum Test Sets}
We also compare our models against a more densely labeled test set of 400 generated images, 200 for each generative model, where labels for each image were collected from 5 different human evaluators. This represents a significantly less noisy evaluation than the binary test sets, as human judgments are not perfectly consistent, and a continuous metric allows us to compare how well our models correlate with human judgment over the full range of image quality.

In total, we recruited 40 human evaluators from Amazon Mechanical Turk and showed each of them 100 images, 50 real and 50 generated, as well as 100 additional images used in HYPE for quality control. Each image received five different evaluations of ``real"=1 or ``fake"=0, and we computed the mean for each image as the realism score, representing the proportion of human evaluators who judged the image as real. For generated images, this can be interpreted as the rate at which humans were ``fooled''.

\textbf{Results.} Our models achieve moderate correlations with human judgments when compared by continuous quality score. Here, the models trained on both datasets achieved correlations of 46\% with human votes, while both models performed worse on the StyleGAN data, correlated by 30\% and 31\% respectively. 




\begin{table}[]
\centering
\caption{Binary accuracy and nonparametric rank-order correlation (Spearman's $\rho$) values compared to human scores for each model and test set combination.}
\label{table:1}
\begin{tabular}{ccccc}
\multicolumn{1}{c}{} & \multicolumn{2}{c}{\textbf{Binary Accuracy}} & \multicolumn{2}{c}{\textbf{Spearman's $\rho$}} \\
\multicolumn{1}{c}{} & \multicolumn{1}{c}{ProGAN Test} & \multicolumn{1}{c}{StyleGAN Test} & \multicolumn{1}{c}{ProGAN Test} & \multicolumn{1}{c}{StyleGAN Test} \\
\hline
\multicolumn{1}{c}{Trained on ProGAN data}   & 65.0\% & 64.7\% & 0.45 & 0.32 \\
\multicolumn{1}{c}{Trained on StyleGAN data} & 58.8\% & 66.4\% & 0.46 & 0.31 \\
\hline
\end{tabular}
\end{table}

\section{Related Work}
A similar approach to encoding human judgment of sample quality into a repeatably applicable neural model has been shown by \citet{lowe2017towards} in the context of natural-language processing. This work was extended by \citet{hashimoto2019unifying} to evaluate the quality of natural-language generative models by combining a neural model for evaluating sample quality with a statistical measure of the diversity of samples generated by a model. 

In the image domain, \citep{misra2016seeing} modeled human reporting bias on sample-level classification labels. Specific to generated images, measuring sample-level quality has been proposed by modifying a precision metric, though alignment with human realism was not explored~\citep{kynkaanniemi2019improved}. Human judgment of the perceptual similarity (as opposed to quality) of images has been predicted by using deep network features by \cite{zhang2018unreasonable}. \cite{prashnani2018pieapp} used deep features to predict human judgments of perceptual image quality, but only in an image-vs-image framing that relies on reference images. \cite{kang2014convolutional} demonstrated automatic evaluations of image quality, but for detecting distortions in real images as opposed to evaluating the realism of generated images. 

Finally, the truncation trick has served many lines of work for trading off fidelity and diversity when sampling from a generator, though it does not necessarily capture human realism~\citep{brock2018large}. Ultimately, most of these metrics in the image domain do not adequately capture human perceptions of realism, meaning the community largely resorts to human-based sample evaluation.

\subsection{Human Evaluation}
Numerous studies have previously conducted human evaluations on the quality of generated images. Several lines of work opt for showing one image at a time~\citep{zhou2019hype, denton2015deep, salimans2016improved, huang2017stacked}. Other works showed real and fake comparisons, asking which of the two is real~\citep{zhang2016colorful,isola2017image,xiao2018generating}. Finally, ranking and triplet comparisons have also been used to sort or select more realistic generated images~\citep{snell2017learning, zhang2017stackgan}.

\subsection{Existing Automated Metrics}
Widely reported metrics Inception Score~\citep{salimans2016improved} and Fr\'echet Inception Distance or FID~\citep{heusel2017gans} fail to correlate well with human perceptual realism, in addition to several other shortcomings, making them easy to optimize without necessarily indicating improvement in models or progress in the field~\citep{barratt2018note, borji2019pros}. However, these automated metrics are on the population level, inspecting statistics on a large (often totalling 100K) distribution of real and fake images, while our measure looks at the sample-level statistics and provides realism scores on individual images. Historically, density estimation is used, but has shown to be inaccurate on high dimensional data~\citep{theis2015note} and, in the generative adversarial network (GAN) training setting, intractable.

\section{Discussion and Conclusion}
In this paper, we introduce a novel method for evaluating sample-level fidelity of generated images that takes human perceptions of realism into account, and experiment with samples from two generative models, ProGAN and StyleGAN. We find that our model's outputs agree between 65.0\%-66.4\% with human-labeled samples on the same model, and 58.8\%-64.7\% on out-of-distribution samples generated by another model. These accuracies match inter-human agreement rates of 58.9-65.5\% on samples from both models. Our work begins to partner statistical approaches with human realism in generative evaluation and to shift automated evaluation towards the aim of rigorously correlating realism metrics with human perception.

While our models' accuracy is far from perfect, they are able to capture a significant amount of signal from images which corresponds to that which humans rely on for evaluating quality. In future work, this realism metric can be easily incorporated with overall diversity metrics, such as in precision and recall. Importantly, these automatic metrics of image quality also generalize across data sets, showing promise for future use on images from novel models and even novel datasets. In this study, we used a deliberately constrained model over a small training set as a proof of concept; collecting more data and training an end-to-end neural model has promise to achieve much more accurate and general results. 

\section*{Acknowledgements}
We thank Michael S. Bernstein and Ranjay Krishna for their helpful discussions throughout the paper. We also appreciate all the Amazon Mechanical Turk workers who participated in our study.

\newpage

\bibliographystyle{icml2018}
\bibliography{main}


\end{document}

%% file: macro.tex
\newtheorem{definition}{Definition}

\newcommand{\bpf}{\begin{proof}}
\newcommand{\epf}{\end{proof}}
\newcommand{\benum}{\begin{enumerate}}
\newcommand{\eenum}{\end{enumerate}}
\newcommand{\bitem}{\begin{itemize}}
\newcommand{\eitem}{\end{itemize}}
\newcommand{\benumalph}{\begin{enumerate}[label=(\alph*)]}
\newcommand{\benumnum}{\begin{enumerate}[label=(\arabic*)]}

\newcommand{\B}{\mathcal{B}}

\newcommand{\D}{\mathcal{D}}

\newcommand{\M}{\mathcal{M}}
\newcommand{\X}{\mathcal{X}}
\renewcommand{\L}{\mathcal{L}}
\newcommand{\U}{\mathcal{U}}
\renewcommand{\P}{\mathcal{P}}

\newcommand{\x}{\textbf{x}}

\newcommand{\z}{\textbf{z}}
\newcommand{\E}{\mathbb{E}}

\newcommand{\argmin}{\text{arg} \min}
\newcommand{\argmax}{\text{arg} \max}

\newcommand{\sumn}{\sum_{i = 1}^n}

\newcommand{\bern}{\text{Bern}}
\renewcommand{\i}{^{(i)}}
\renewcommand{\t}{^{(t)}}

\newcommand*{\newlinecommand}[2]{%
  \newcommand*{#1}{%
    \begingroup%
    \escapechar=`\\%
    \catcode\endlinechar=\active%
    \csname\string#1\endcsname%
  }%
  \begingroup%
  \escapechar=`\\%
  \lccode`\~=\endlinechar%
  \lowercase{%
    \expandafter\endgroup
    \expandafter\def\csname\string#1\endcsname##1~%
  }{\endgroup#2\space}%
}

\newlinecommand{\hh}{\textbf{#1}}
\newcommand{\tocite}{\textbf{\textcolor{orange}{(TO CITE)}} }

\DeclarePairedDelimiter\abs{\lvert}{\rvert}
\DeclarePairedDelimiter\norm{\lVert}{\rVert}
\makeatletter
\let\oldabs\abs
\def\abs{\@ifstar{\oldabs}{\oldabs*}}
\let\oldnorm\norm
\def\norm{\@ifstar{\oldnorm}{\oldnorm*}}
\makeatother

\newcommand{\Q}{\mathbb{Q}}

\renewcommand{\P}{\mathbb{P}}

\setenumerate{noitemsep}
\setitemize{noitemsep}
\setitemize{topsep=-1mm}